\pdfoutput=1
\documentclass[10pt]{article}

\usepackage{styles/arxiv}      

\usepackage{graphicx}
\usepackage{booktabs}
\usepackage{multirow}
\usepackage{array}
\usepackage{pgfplots}
\pgfplotsset{compat=1.18}
\usepackage[font=small]{caption}
\usepackage[nospace,nocompress,nobreak]{cite}  

\usepackage{packages/math}     
\usepackage{packages/refs}     
\usepackage{packages/symbols}

\title{Constructing Reverse Thinking: Developing Large Language Models' Reverse Thinking Ability}

\author{Xin Liu\textsuperscript{1}, Yunhai Li\textsuperscript{1}, Chunfu Jia\textsuperscript{3}, Ziliang Chen\textsuperscript{2}, and Jisen Song\textsuperscript{1}\\[4pt]
\small\textsuperscript{1}China University of Petroleum (East China), Qingdao, China\\
\small E-mail: lx@upc.edu.cn, liyunhai2025@163.com, sdqdjmsjs@163.com\\[3pt]
\small\textsuperscript{2}Goertek Inc., Qingdao, China\\
\small E-mail: z23070047@s.upc.edu.cn\\[3pt]
\small\textsuperscript{3}College of Cryptology and Cyber Science, Nankai University, Tianjin, China\\
\small E-mail: cfjia@nankai.edu.cn}

\date{}

\begin{document}
\maketitle

\begin{abstract}
When facing complex problems, humans tend to try various ideas for different issues. Human thinking patterns exhibit remarkable flexibility in adapting to diverse scenarios. GPT-o1 and GPT-o3 released by OpenAI, and DeepSeek-R1 released by DeepSeek adopt long chain-of-thought models to address complex problems by increasing reasoning depth. However, they default to a forward reasoning mode that starts from the problem and initial conditions. We conducted statistical analysis on the accuracy of different mathematical problem datasets on models of different scales, and found that there are five reasons for errors: Insufficient solution-space coverage, Computational mistakes, Unverified assumptions, Ignoring constraint conditions, Maximum response length limitation. To address the above issues, we propose a backward reasoning pattern construction method aimed at enhancing the model's reverse thinking ability and dynamic adaptability when handling complex problems. First, we constructed an easy-hard two-stage Math dataset for training large models and gradually improving the inference ability of the large model at different difficulty levels. A general large model is applied to construct backward reasoning paths for complex problems, which are then integrated with forward reasoning paths to create a balanced training dataset covering both reasoning paradigms. The dataset is further divided into easy and hard subsets based on difficulty. And a two-stage supervised fine-tuning process is applied to progressively train the model's backward reasoning capability from shallow to deep. Furthermore, a fine-grained reward mechanism is developed, employing smoothed reward signals to strengthen the model's ability to autonomously select thinking modes during the reasoning process, thereby avoiding reward hacking. A linear-decay balanced sampling strategy is designed to maintain a balance between forward and backward reasoning path samples during training, enabling the model to converge quickly and stably. Experimental results show that our method significantly improves reasoning efficiency and accuracy in tasks such as mathematical proofs, offering a flexible and efficient reasoning paradigm for solving complex problems.
\end{abstract}


\section{Introduction}

Humans solve problems not by following a single fixed thinking path, but through a dynamic combination of two basic modes: forward thinking and reverse thinking. Typically, people prefer to reason forward from known conditions---this direct and efficient approach dominates everyday cognition. However, when problems become complex or the forward path is obstructed, humans actively switch to reverse thinking: backtracking from conclusions to conditions, retracing steps from goals, or assuming the opposite to seek contradictions. Effective problem-solving frequently involves the flexible integration of forward and reverse thinking.

Taking a geometry proof as an example: in triangle $ABC$, it is given that $\triangle ABC$ is not isosceles (i.e., $AB \neq AC$), and $AD$ is the bisector of $\angle BAC$. To prove that $AD$ is not perpendicular to $BC$ by contradiction, we start by assuming the opposite: $AD \perp BC$. Since $AD$ is the angle bisector of $\angle BAC$, we have $\angle BAD = \angle CAD$. From the assumption $AD \perp BC$, we get $\angle ADB = \angle ADC = 90^{\circ}$. Now consider triangles $\triangle ABD$ and $\triangle ACD$: they share side $AD$, and we have $\angle BAD = \angle CAD$ and $\angle ADB = \angle ADC$. By the Angle-Side-Angle (ASA) congruence criterion, $\triangle ABD \cong \triangle ACD$. Therefore, the corresponding sides $AB$ and $AC$ are equal, i.e., $AB = AC$. But this contradicts the given condition that $AB \neq AC$. Hence, our assumption that $AD \perp BC$ must be false. Therefore, $AD$ is not perpendicular to $BC$.

Reverse thinking problems typically require long chain-of-thought reasoning, as they demand step-by-step backward tracing from conclusions to conditions. To investigate how existing long chain-of-thought models perform across different types of reasoning problems, we conducted a statistical analysis of the problem categories in three mathematical test sets: GSM8K, MATH-500, and AIME. GSM8K primarily focuses on elementary and middle-school mathematics, MATH-500 contains more complex, higher-level mathematical problems, and AIME consists of intermediate to advanced math competition questions. Across these three datasets, we identified 171 problems that can be solved more efficiently using backward reasoning (referred to as ``backward questions''). On these backward questions, smaller open-source models perform poorly and often fail to complete the tasks within the prescribed maximum response length. In contrast, larger closed-source models, such as GPT-5, show improved accuracy but still exhibit a certain percentage of failure cases.

\begin{figure}[t]
  \centering
  \resizebox{\linewidth}{!}{%
  \begin{tikzpicture}
  \begin{axis}[
      width=\linewidth, height=6.2cm,
      ybar, bar width=2.6pt,
      ymin=0, ymax=0.20,
      axis y line*=left,
      ylabel={Percentage of errors},
      symbolic x coords={Qwen3-1.7b,Deepseek-R1-1.5b,Qwen3-8b,Deepseek-R1-7b,Qwen3-14b,Deepseek-R1-14b,Qwen3-32b,Deepseek-R1-32b,GPT-5,GPT-o3},
      xtick=data,
      x tick label style={rotate=35,anchor=east,font=\scriptsize},
      legend style={font=\tiny,at={(0.5,-0.42)},anchor=north,legend columns=3},
      tick label style={font=\scriptsize},
    ]
    \addplot coordinates {(Qwen3-1.7b,0.0217) (Deepseek-R1-1.5b,0.0435) (Qwen3-8b,0) (Deepseek-R1-7b,0.0326) (Qwen3-14b,0) (Deepseek-R1-14b,0.0217) (Qwen3-32b,0) (Deepseek-R1-32b,0) (GPT-5,0.0109) (GPT-o3,0.012)};
    \addplot coordinates {(Qwen3-1.7b,0.0326) (Deepseek-R1-1.5b,0.0543) (Qwen3-8b,0.0109) (Deepseek-R1-7b,0.0109) (Qwen3-14b,0) (Deepseek-R1-14b,0.0109) (Qwen3-32b,0.0109) (Deepseek-R1-32b,0) (GPT-5,0) (GPT-o3,0.012)};
    \addplot coordinates {(Qwen3-1.7b,0.0435) (Deepseek-R1-1.5b,0.0543) (Qwen3-8b,0.0109) (Deepseek-R1-7b,0.0217) (Qwen3-14b,0.0109) (Deepseek-R1-14b,0.0652) (Qwen3-32b,0.0109) (Deepseek-R1-32b,0.0109) (GPT-5,0.0217) (GPT-o3,0.025)};
    \addplot coordinates {(Qwen3-1.7b,0.0543) (Deepseek-R1-1.5b,0.0326) (Qwen3-8b,0) (Deepseek-R1-7b,0.0109) (Qwen3-14b,0.0109) (Deepseek-R1-14b,0.0109) (Qwen3-32b,0) (Deepseek-R1-32b,0.0109) (GPT-5,0.0217) (GPT-o3,0.006)};
    \addplot coordinates {(Qwen3-1.7b,0.1413) (Deepseek-R1-1.5b,0.1522) (Qwen3-8b,0.1087) (Deepseek-R1-7b,0.087) (Qwen3-14b,0.0978) (Deepseek-R1-14b,0.0761) (Qwen3-32b,0.0326) (Deepseek-R1-32b,0.0543) (GPT-5,0) (GPT-o3,0)};
    \legend{(a) Insufficient solution-space coverage,(b) Computational mistakes,(c) Unverified assumptions,(d) Ignoring constraints,(e) Max response length}
  \end{axis}
  \begin{axis}[
      width=\linewidth, height=6.2cm,
      ymin=0, ymax=9000,
      axis y line*=right, axis x line=none,
      ylabel={Average response length},
      symbolic x coords={Qwen3-1.7b,Deepseek-R1-1.5b,Qwen3-8b,Deepseek-R1-7b,Qwen3-14b,Deepseek-R1-14b,Qwen3-32b,Deepseek-R1-32b,GPT-5,GPT-o3},
      tick label style={font=\scriptsize},
    ]
    \addplot+[mark=*,no markers,thick,black] coordinates {(Qwen3-1.7b,7892) (Deepseek-R1-1.5b,6902) (Qwen3-8b,7290) (Deepseek-R1-7b,6345) (Qwen3-14b,6981) (Deepseek-R1-14b,5587) (Qwen3-32b,6123) (Deepseek-R1-32b,5318) (GPT-5,5123) (GPT-o3,926)};
  \end{axis}
  \end{tikzpicture}%
  }
  \caption{Error type distribution and average response length statistics.}
  \label{fig:error}
\end{figure}
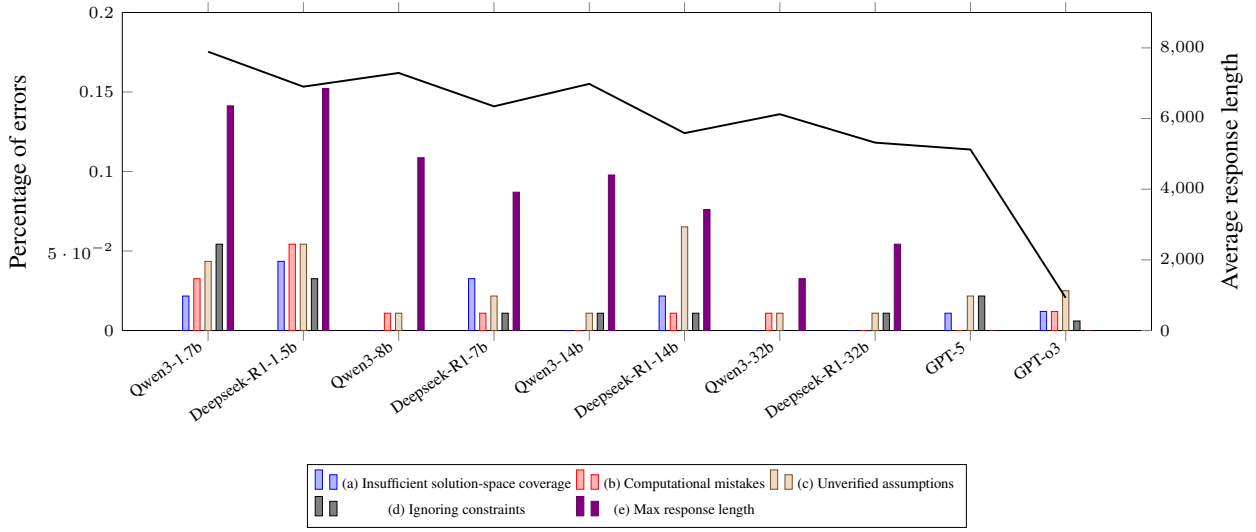

We conducted an in-depth analysis of the failure cases in these backward questions and found that the errors can be broadly classified into five categories, as shown in Fig.~\ref{fig:error}.

\begin{itemize}
  \item[(a)] \textbf{Insufficient solution-space coverage:} The model fails to consider all possible cases, resulting in missing valid solutions. For example, in combinatorial problems, it may neglect certain permutations or combinations.
  \item[(b)] \textbf{Computational mistakes:} Errors occur in numerical calculations or formula simplifications, such as when solving equations or manipulating fractions.
  \item[(c)] \textbf{Unverified assumptions:} To simplify the problem or converge quickly to an answer, the model asserts certain conditions as necessarily true without rigorous validation, causing all subsequent reasoning to rest on a false premise. An example is wrongly assuming two line segments are equal in a geometric proof.
  \item[(d)] \textbf{Ignoring constraint conditions:} In many tasks, the answer is subject to constraints (e.g., the result must be an integer or lie within a specific range). The model overlooks these restrictions during reasoning, leading to the exploration of numerous invalid solution paths.
  \item[(e)] \textbf{Maximum response length limitation:} The model fails to complete the problem within the specified maximum output length (e.g., 16K tokens). On average, larger models produce shorter responses than smaller ones. This is because smaller models tend to generate more redundant or incorrect content during inference, which increases the length of their outputs.
\end{itemize}

As shown in Fig.~\ref{fig:error}, smaller models exhibit all categories of errors, whereas larger commercial models do not present obvious computational mistakes or length-limit issues. However, they still suffer from insufficient solution-space coverage, unverified assumptions, and neglect of constraints, indicating that these are common challenges for model reasoning capabilities.

The failure patterns we observed reflect a common underlying issue: difficulty in guiding forward search toward the correct solution. Reverse thinking offers a complementary strategy. When a problem provides a clear goal and its reasoning steps are reversible, backward reasoning can use the goal to constrain each step, eliminating irrelevant branches and reducing the need for speculative assumptions. Therefore, we developed a method for training large language models to adopt reverse thinking, aiming to improve their accuracy on the aforementioned backward questions.

Long Chain-of-Thought (Long CoT) reasoning is categorized into three main types based on reasoning structure: Deep Reasoning, Extensive Exploration, and Feasible Reflection~\cite{ref1}.

Deep Reasoning aims to overcome the limitation of Short CoT on the number of logical nodes, enabling deep, multi-step analysis. It mainly comprises two types of methods. The first is deep reasoning formats, including natural language reasoning \cite{ref2,ref3}, structured language reasoning such as code/symbolic language \cite{ref4}, and latent space reasoning \cite{ref5}. The second is deep reasoning learning, which includes imitation learning from advanced models or human data \cite{ref6}, as well as self-learning through preference optimization or reinforcement learning \cite{ref7,ref8}. Recent work has also revealed the alternating pattern of exploration and exploitation at the neuron activity level \cite{ref9}. The core challenge of deep reasoning is its limited generalization ability. Rameshkumar et al. demonstrated that as the complexity of reasoning problems exceeds the training distribution, the performance of large reasoning models degrades sharply, and they fail to generalize across varying levels of complexity \cite{ref10}. This suggests that current deep reasoning capabilities reflect memorization of training patterns rather than genuine logical understanding.

Extensive exploration aims to enhance performance on uncertain problems by increasing the diversity of reasoning paths. Structurally, these methods include tree search \cite{ref11,ref12}, beam search \cite{ref13,ref14}, and Monte Carlo Tree Search \cite{ref15,ref16,ref17}. Reasoning diversity can also be improved through sequential scaling \cite{ref18} or parallel sampling \cite{ref19}. Coupling language models with strategic search reached human-level play in the game of Diplomacy \cite{ref20}, and combining large language models with evolutionary program search has led to new mathematical discoveries \cite{ref21}. However, such exploration often leads to overthinking \cite{ref22}. Chen et al. \cite{ref23} identified an optimal trade-off between reasoning length and accuracy, motivating later work on early exit, chain-of-thought compression, and adaptive reasoning \cite{ref24} to mitigate this issue. The primary bottleneck of extensive exploration lies in its substantial computational overhead. For instance, on the 24-point task, Tree-of-Thoughts (ToT) incurs roughly two orders of magnitude higher inference cost than Chain-of-Thought (CoT). While dynamic scaling strategies offer partial relief, striking a favorable balance between exploration breadth and computational efficiency remains an open challenge. Moreover, most existing tree- or graph-based reasoning frameworks rely on pre-defined, static reasoning structures, which limits their adaptability to dynamic or unseen problem types.

Feasible Reflection enables models to self-correct through feedback and refinement mechanisms. This process consists of two stages. The feedback stage employs overall feedback such as Outcome Reward Models \cite{ref25}, process feedback such as Process Reward Models \cite{ref26}, or a combination of both \cite{ref27}. The refinement stage optimizes reasoning paths through prompt-based iterative generation \cite{ref28}, error correction imitation under supervised fine-tuning \cite{ref29}, and reinforcement learning-driven self-correction \cite{ref30}. Recent advances show that the RL-zero training paradigm can internalize self-reflection and self-correction as inherent cognitive capabilities of the model \cite{ref31}, while structured multi-perspective evaluation further improves the robustness of self-correction \cite{ref32}. The core challenge of feasible reflection is the trade-off between iteration cost and effectiveness. Each round of reflection requires additional model calls, leading to linear growth in inference latency and computational cost. More critically, the effectiveness of reflection is highly dependent on the capability of the base model---weaker models struggle to generate effective feedback and may fall into a situation where ``more reflection leads to more errors.''

The above work is dedicated to obtaining correct answers to complex problems, but no training has been conducted on reverse thinking of large models. Justin Chih Yao Chen et al. proposed Reverse Enhanced Thinking (REvThink) \cite{ref33} to train the reverse reasoning ability of student model by learning bidirectional reasoning paths. While its effectiveness is inherently bounded by the student model's forward reasoning capacity, as reverse reasoning cannot inject new knowledge but only regularizes existing reasoning paths. Moreover, their data filtering strategy retains only teacher-validated correct samples, which may exclude hard cases critical for out-of-distribution generalization. Finally, the training-inference discrepancy (exposure bias) remains an unresolved issue.

To address the above challenges, we introduce a method for constructing a backward reasoning paradigm. By training the model to enhance its backward reasoning ability, this approach mitigates issues often encountered in forward reasoning, such as ``insufficient solution-space coverage'', ``ignoring constraints'', and ``unverified assumptions''. Training a model's backward reasoning ability requires datasets annotated with backward reasoning processes. We first constructed a dataset containing both forward and backward reasoning processes, and divided the dataset into ``easy'' and ``hard'' categories based on problem difficulty we called Easy-hard split. We dynamically adjusted the sampling ratio of forward and backward reasoning examples to perform dual-stage supervised fine-tuning. Furthermore, we design a fine-grained reward function to encourage the model to flexibly select the optimal reasoning path in appropriate task scenarios. We not only validated the enhancement of reverse thinking ability through training long-chain thinking models, but also demonstrated the potential of backward reasoning in improving inference efficiency and accuracy.

\section{Related Work}

\subsection{Prompt-Engineering Driven Reasoning Methods}

Prompt engineering is one of the early research directions aimed at enhancing the reasoning abilities of large models. By adding prompts such as ``let's think step by step'' to form a chain of thought, it improves model performance on arithmetic and logical reasoning tasks. Based on this, Wang et al. proposed sampling multiple reasoning paths simultaneously and selecting the most consistent option in the final answer, thereby improving the accuracy of large models on arithmetic and commonsense reasoning tasks \cite{ref19}. Subsequently, researchers further extended reasoning paths and proposed more sophisticated methods. For example, another representative method is Tree-of-Thought (ToT) \cite{ref11}, which expands reasoning paths into a tree structure and uses search strategies to find the optimal solution. Graph-of-Thought (GoT) \cite{ref12} represents the reasoning process as a graph and leverages aggregation, refinement, and generation operations to improve performance on complex problems. Another category of tasks adopts task decomposition, which breaks a complex problem into simpler subproblems and solves them either sequentially or in parallel to mitigate reasoning complexity \cite{ref34,ref35,ref36}. Building on this, Chen et al. introduced ``Boosting of Thoughts,'' which leverages iterative exploration and self-evaluation to accumulate trial-and-error across reasoning trees, dynamically refining prompts to strengthen complex reasoning \cite{ref37}. Although the above methods improve the reasoning capabilities of large language models to some extent, they often require meticulously designed prompts, are sensitive to phrasing, and do not transfer well across tasks. While prompting can elicit existing reasoning and boost performance, it invokes rather than extends reasoning capacity. In some pattern-based in-context learning tasks, Chain-of-Thought prompting may even degrade performance by disrupting implicit reasoning \cite{ref2}. This indicates that explicit solution paths can disrupt implicit reasoning mechanisms, thereby hurting rather than helping performance. Additionally, exploring multiple reasoning paths incurs significant time and token costs.

\subsection{Long Chain-of-Thought Models}

Beyond relying on prompt engineering, recently emerging Long Chain-of-Thought models, such as GPT-o3, Deepseek-R1, kimi-k1.5 \cite{ref38}, and QwQ, use reinforcement learning to extend the model's reasoning length, enabling the formation of longer chains of thought before generating answers, thereby significantly improving reasoning performance without manually designed prompts. However, solely relying on lengthening the chain of thought to improve reasoning performance inevitably leads to excessive consumption of time and other resources, as well as the problem of ``overthinking,'' where the model repeatedly reasons even on simple problems, greatly increasing the waiting cost. To address this problem, Qwen3 employs ``Thinking Mode Fusion'' to unify ordinary conversation and extended chain-of-thought reasoning within the same model, but human intervention is still necessary to decide whether to activate long-chain reasoning \cite{ref39}. Based on this, a series of methods aim to enhance the model's autonomy in switching reasoning modes. For instance, Tu et al. proposed the AutoThink framework, which uses a multi-stage reinforcement learning strategy to train the model to dynamically switch between thinking and non-thinking modes, significantly reducing unnecessary reasoning overhead \cite{ref40}. Jiang et al. proposed Large Hybrid-Reasoning Models (LHRMs), which incorporate hybrid fine-tuning and group-strategy optimization, enabling the model to adaptively decide whether to engage in deep reasoning based on the contextual information of user queries \cite{ref41}. Zhang et al. proposed the Adaptive Self-Recovery Reasoning (ASSR) framework, which introduces an accuracy-aware length reward adjustment mechanism to suppress unnecessary lengthy reasoning \cite{ref42}. The AdaptThink algorithm employs constrained optimization objectives and an importance-sampling strategy, enabling the model to adaptively select between thinking and non-thinking modes based on problem difficulty \cite{ref43}. In summary, although existing research has made significant progress in adaptive reasoning strategies, it mainly focuses on adaptive switching between thinking and non-thinking modes, and has not yet introduced richer reasoning paradigms, such as dynamic selection between forward and backward reasoning.

\subsection{Backward Reasoning}

Some other studies attempt to incorporate the concept of backward reasoning at the prompt-engineering level. Li et al. proposed a hypothesis-testing prompt, which introduces conclusion assumptions and backward verification to enhance the performance of large language models on deductive reasoning tasks \cite{ref44}. Huang et al. proposed RFF, which alternates between backward and forward reasoning to improve reasoning efficiency and accuracy \cite{ref45}. Some researchers have designed backward-reasoning prompts to verify the correctness of multiple candidate answers \cite{ref46,ref47}. Some researchers have also applied backward reasoning to planning scenarios. For instance, Zhang et al. proposed enhancing planning capability in multi-API combination scenarios by using backward reasoning along with stepwise API parameter filling \cite{ref48}. Ren et al. introduced a backward planning approach based on problem reversal, improving large models' performance on graph planning and similar tasks \cite{ref49}. Additionally, some studies have leveraged the concept of backward reasoning for data augmentation. For example, Wen et al. constructed both forward and backward reasoning processes for code vulnerabilities, ensuring the synthesis of high-quality reasoning data \cite{ref50}. Other researchers have added backward questions to mathematical datasets to increase the diversity of training data, enhancing the model's performance in mathematical reasoning \cite{ref51,ref33}. Most of the methods mentioned above remain at the level of prompt design or data augmentation and do not systematically instill backward reasoning capabilities during model training. The method proposed in this paper not only enhances the model's backward reasoning ability but also leverages reinforcement learning to enable the model to select reasoning paths adaptively across different problem types.

\section{Methodology}

\begin{figure}[t]
  \centering
  \includegraphics[width=\linewidth]{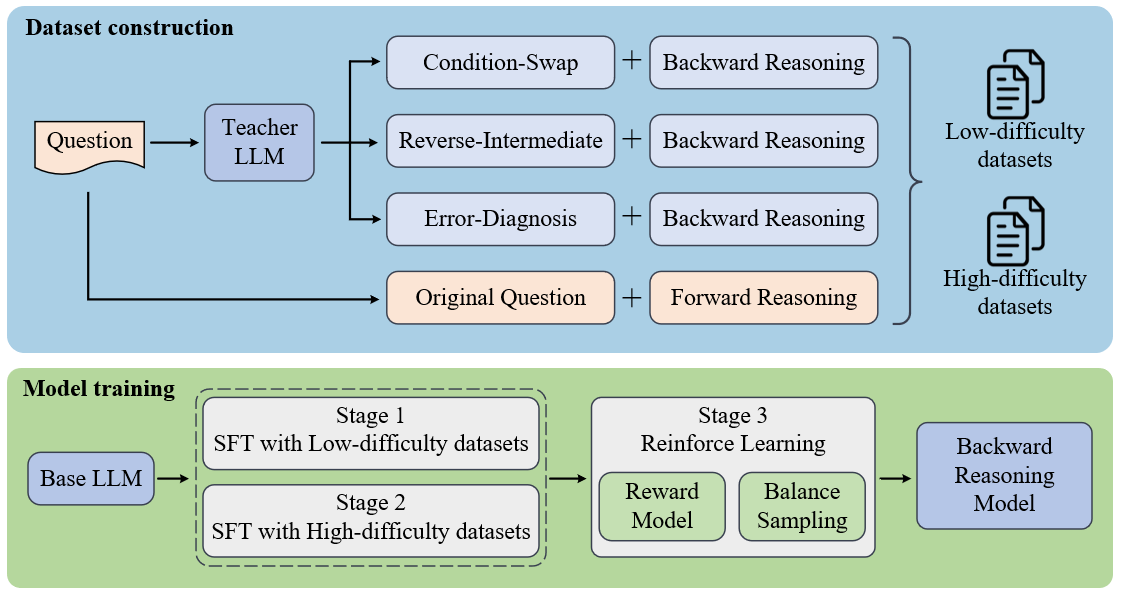}
  \caption{Method framework diagram.}
  \label{fig:framework}
\end{figure}

In this section, we elaborate the construction process for the model's backward reasoning mode, with the overall framework illustrated in Fig.~\ref{fig:framework}. We first constructed two backward reasoning datasets, expanding backward-reasoning samples by generating three types of backward questions through rewriting the original questions. We then performed easy-to-hard dual-stage supervised fine-tuning on those datasets, enabling the model to learn backward-thinking skills based on backward-reasoning trajectories. Secondly, we tailored the training strategy of the Reinforce++ algorithm by designing reward functions for reasoning mode label accuracy, answer correctness, and format compliance. We also introduced a linearly decaying balanced sampling strategy to balance the initial sample distribution, further optimizing the model to flexibly utilize backward-reasoning capabilities.

\subsection{Construction of the Backward-Reasoning Dataset}

Due to the unique value of mathematical reasoning tasks in evaluating a model's logical rigor, step decomposition ability, and application of reverse thinking, we selected datasets used for mathematical reasoning tasks as the main component of our backward reasoning dataset, dividing them into ``hard'' and ``easy'' subsets based on difficulty. We then designed prompts to leverage the Deepseek-R1 large model to generate backward questions and their reasoning paths, forming backward reasoning data comprising the problem and its reasoning trajectory.

We first filtered the MATH-lighteval and OpenR1-Math-220k datasets. In MATH-lighteval, each sample is classified into five levels based on the complexity of solving the problem. We selected samples with difficulty levels below 4 (Level $< 4$) to form an easy-question dataset for the first stage of model training. High difficulty samples with Level $= 5$ were included in a hard-question dataset for the second stage of training, aiming to enhance the model's ability to handle complex reasoning chains.

As for the OpenR1-Math-220k, it contains 220,000 math problems, each of which has two to four reasoning trajectories generated by DeepSeek-R1 for the problems in NuminaMath 1.5. We consider that the length of a model's reasoning path reflects the problem's difficulty. Therefore, we filtered the Default subset of OpenR1-Math (approximately 94,000 problems) based on the token length of reasoning path, retaining the top 10\% longest trajectories to form a high complexity subset. This subset, together with Level 5 samples from MATH-lighteval, forms the hard-question dataset for the second stage of training.

For the two datasets, we designed prompt templates and applied Deepseek-R1 to generate high-quality samples with backward reasoning paths. First, we constructed a prompt template PromptBQ for generating backward questions. This template takes the original problem and its forward-solving steps as input context to produce three types of backward questions: (1) Condition-swap: using the answer or key conclusion of the original problem as new known information, while treating some of the original conditions as unknown, thereby forming a new problem; (2) Reverse-intermediate: selecting an intermediate result from the original forward solution steps and requiring the model to infer the necessary premises that lead to this result, thereby constructing a complete backward reasoning task; (3) Error-diagnosis: provided with the original problem and its forward solution steps, the model must alter the solution to include at least one incorrect step, thereby creating an error-diagnosis task. This process can be formally represented as:

\begin{equation}
\{Q_{\mathrm{swap}}, Q_{\mathrm{int}}, Q_{\mathrm{err}}\} = f(Q_{\mathrm{ori}}, R_{\mathrm{for}})
\label{eq:generate}
\end{equation}

\noindent Where $Q_{\mathrm{ori}}$ represents the original problem, $R_{\mathrm{for}}$ represents the problem-solving steps of forward reasoning for the original problem, and $Q_{\mathrm{swap}}$, $Q_{\mathrm{int}}$, and $Q_{\mathrm{err}}$ represent three types of backward questions: condition-swap, reverse-intermediate, and error-diagnosis respectively.

The specific design of the PromptBQ is as follows:

\begin{quote}\small
``Given a math problem and its correct answer, create 3 mathematically valid inverse problems.

Original Problem:

\{input\_question\}

Correct solution: \verb|\boxed{{correct_answer}}|

Requirements:

- Each problem must be uniquely solvable and logically linked to the original.

- Keep difficulty and topic similar.

- No solving, only output problems, the problem must be described clearly.

- Do not use \verb|\boxed{}| in your problems.

Types:

1) Condition-Swap: Swap knowns and unknowns so the new target is solvable.

2) Reverse-Intermediate: Give a valid intermediate result from the original solution, and infer the necessary conditions that ensure the validity of this intermediate step.

3) Error-Diagnosis: Alter the original solution to generate a solution containing at least one error, without identifying the erroneous steps.

Output exactly 3 items in this format:

Type: <type>

Problem: <problem statement>''
\end{quote}

After adding the backward question, a backward reasoning prompt template PromptBR needs to be designed. The backward question is input into the model, causing it to generate the corresponding backward reasoning path. The PromptBR emphasizes that the model must follow a complete and verifiable reasoning chain, starting from the target state and gradually backtracking. It also requires the output to begin with the \texttt{<backward>} tag to explicitly annotate the backward reasoning path. The final prompt template adopted after iterative optimization is as follows:

\begin{quote}\small
``You are a mathematical reasoning assistant. Please use backward reasoning to solve the following problem. Backward reasoning means that you should start from the final target state, gradually backtrack on the intermediate conditions it depends on, until reaching the initial known conditions, and build a complete logical chain.

Put the complete reasoning chain within the <backward> and </backward>

Problem:\{input\_question\}''
\end{quote}

Based on PromptBQ, a corresponding backward question is generated for each original question. Then, based on PromptBR, Deepseek-R1 is used to generate backward reasoning paths. For multiple solutions to each question, we compared their efficiency and accuracy. When all answers are correct, to reduce redundant reasoning burden during training, we prioritized retaining the solution with the lowest token cost. To prevent the model from catastrophically forgetting past forward reasoning patterns during learning, we retained a certain proportion of forward reasoning samples in the dataset.

To ensure the reliability and logical consistency of the newly expanded backward questions and their reasoning paths, for each generated backward question, we re-input the question and the reasoning process into DeepSeek-R1, asking it to determine whether the answer makes the question true and to provide a binary classification judgment. Through the above process, we finally constructed two backward reasoning datasets: Datasets1 (low to medium difficulty) and Datasets2 (high difficulty).

\subsection{Dual-stage supervised fine-tuning}

We choose Deepseek-R1-distill-1.5b, Qwen3-1.7b, Qwen3-8b, and Qwen3-14b as the base models, and perform cold-start fine-tuning using the Low-Rank Adaptation (LoRA) algorithm. With all original base model parameters kept frozen, only a small set of newly introduced trainable weights is updated, enabling efficient and lightweight fine-tuning. To reduce the difficulty of learning backward reasoning tasks and ensure training stability, we adopt a dual-stage supervised fine-tuning strategy that enables the model to gradually acquire backward reasoning from simple to complex cases. In the first stage, Datasets1 is used for training. Its problems have clear structures and short reasoning chains, allowing the model to focus on establishing fundamental backward reasoning patterns. In the second stage, the more complex Datasets2 is introduced to enlarge the search space and increase reasoning depth, thereby further strengthening and expanding the model's backward reasoning capabilities to handle more diverse and complex scenarios. This progressive training strategy effectively prevents the model from falling into local optima too early when handling complex tasks, thereby achieving more stable convergence and higher reasoning accuracy.

\subsection{Reinforcement learning optimization}

We expect the model to adaptively leverage backward reasoning when necessary, while still retaining and utilizing forward reasoning instead of abandoning it entirely. We extend the training strategy of the Reinforce++ algorithm by introducing a balanced sampling strategy with linear decay, further enhancing the model's ability to select between forward and backward reasoning modes.

In the previous section, we expanded the backward-reasoning samples and filtered them based on accuracy and average response length. We then assigned forward and backward labels, treating the task as a binary classification problem, and trained a discriminator reward model based on BERT to determine whether a given problem is better suited for forward or backward reasoning. This reward model is used to assign rewards to the first token (\texttt{<think>} or \texttt{<backward>}).

To prevent reward hacking during reinforcement learning---such as skipping the reasoning process and directly giving an answer, making random guesses, or providing irrelevant content to evade reasoning---we design a fine-grained reward signal. The reward function considers both output format and answer correctness, emphasizing the reasoning process and the accuracy of the final response.

The model's output must strictly comply with the specified format, using regular expressions to determine whether the output contains the \texttt{<think>} or \texttt{<backward>} tags and whether a reasoning process is present within these tags. The formula for calculating the format reward is given below:

\begin{equation}
R_{\mathrm{format}} =
\begin{cases}
R_{m}, & \mathrm{token_{first}} \in \{\mathrm{token}_{f}, \mathrm{token}_{b}\}\\[2pt]
0, & \mathrm{otherwise}
\end{cases}
\label{eq:format}
\end{equation}

\noindent Here, $\mathrm{token_{first}}$ represents the first token generated, $\mathrm{token}_{f}$ and $\mathrm{token}_{b}$ represent \texttt{<think>} and \texttt{<backward>} respectively. When the first token belongs to the \texttt{<think>} or \texttt{<backward>} tag, the score is calculated using the reward model $R_{m}$.

After format verification, we evaluate the correctness of the final answer. Meanwhile, we impose constraints on the length of the chain of thought, as excessively long reasoning greatly increases latency. For reasoning that yields a correct answer, we prefer shorter chains of thought; for reasoning that leads to an incorrect answer, we impose a larger penalty on shorter chains, encouraging the model to extend its reasoning to explore more branches. Since model response length typically falls within a fixed interval (from 0 to the maximum length), we introduce a cosine function to smooth the reward score. The computation formula for the answer reward is as follows:

\begin{equation}
R_{\mathrm{answer}} =
\begin{cases}
\cos\!\left(\dfrac{\pi L}{2 L_{\max}}\right), & \mathrm{correct}\\[8pt]
-\cos\!\left(\dfrac{\pi L}{2 L_{\max}}\right), & \mathrm{incorrect}
\end{cases}
\label{eq:answer}
\end{equation}

\noindent Here, $L$ denotes the length of the model-generated content, and $L_{\max}$ represents the predefined maximum response length. The cosine function $\cos$ is used to smooth the reward based on the response length.

The final reward score is computed as a weighted sum of the format reward and the answer reward.

\begin{equation}
R = \alpha\, R_{\mathrm{format}} + \beta\, R_{\mathrm{answer}}
\label{eq:reward}
\end{equation}

\begin{figure}[t]
  \centering
  \includegraphics[width=0.9\linewidth]{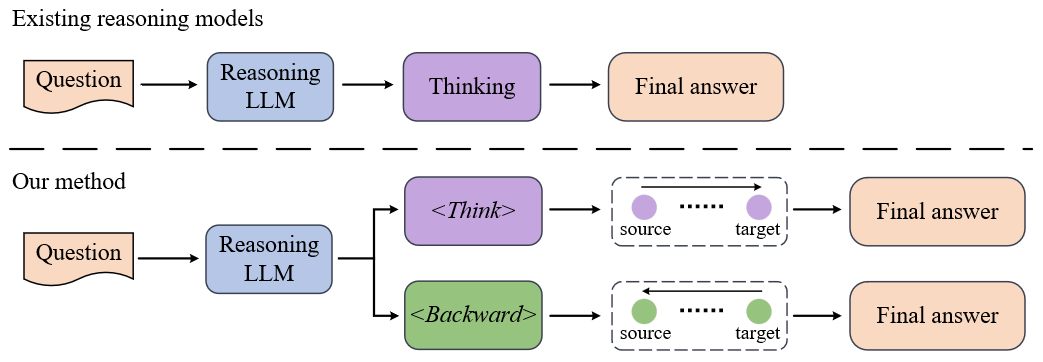}
  \caption{Comparison of model reasoning methods.}
  \label{fig:methods}
\end{figure}

To enable the model to extensively explore both forward and backward reasoning in the early stages of training, we introduce a linearly decaying balanced sampling strategy, ensuring that the model receives an equal proportion of forward and backward samples at the beginning of training.

\begin{equation}
n_{\mathrm{assign}}(t) = w(t)\cdot N
\label{eq:sampling}
\end{equation}

\noindent Where $N$ represents the total number of samples, $n$ represents the current number of samples, and $w(t)$ represents the decay coefficient, which is calculated as follows:

\begin{equation}
w(t) = \max\!\left(0,\; 1 - \frac{t}{T}\right)
\label{eq:decay}
\end{equation}

\noindent By introducing a decay coefficient, some samples within a training batch are explicitly assigned a reasoning mode, guaranteeing balanced exploration of both reasoning paths at the start of training. As training steps increase, this explicit assignment diminishes linearly, decreasing the number of samples constrained to a specific reasoning mode, and the model progressively begins to select the suitable reasoning mode autonomously.

We adopt Reinforce++ as our reinforcement learning algorithm, directly optimizing the policy gradient to maximize the expected cumulative reward. The objective function is given as follows:

\begin{equation}
\mathcal{J}(\theta) = \mathbb{E}\!\left[\min\!\left(r_{t}(\theta) A_{t},\; \mathrm{clip}\!\left(r_{t}(\theta), 1-\epsilon, 1+\epsilon\right) A_{t}\right)\right]
\label{eq:ppo}
\end{equation}

\noindent Here, $r_{t}(\theta)$ is the policy ratio, which is the ratio of the action probability under the new policy $\pi_{\theta}$ to that under the old policy $\pi_{BL}$ at time step $t$. The clipping function $\mathrm{clip}$ limits the policy ratio within a predetermined range $[1-\epsilon, 1+\epsilon]$. $A_{t}$ represents the advantage function, calculated as follows:

\begin{equation}
A_{t} = r(x,y) - \beta \sum_{i} \mathrm{KL}(i)
\label{eq:adv}
\end{equation}

\noindent Here, $r(x,y)$ is the reward function, and $\mathrm{KL}(i)$ is the KL divergence estimate for each token, used to penalize deviations from the initial model, ensuring that the model does not stray excessively from the original distribution while optimizing rewards. The KL divergence is computed using the same method as in PPO.

\begin{equation}
\mathrm{KL}(i) = \frac{\pi_{\theta}(o_{i}\mid q, o_{<i})}{\pi_{BL}(o_{i}\mid q, o_{<i})} - \log\frac{\pi_{\theta}(o_{i}\mid q, o_{<i})}{\pi_{BL}(o_{i}\mid q, o_{<i})} - 1
\label{eq:kl}
\end{equation}

As shown in Fig.~\ref{fig:methods}, compared with standard reasoning models, our method can automatically select the appropriate reasoning mode upon receiving a question. The detailed experimental configuration will be presented in Section~\ref{sec:experiments}.

\section{Experiments}
\label{sec:experiments}

\subsection{Experimental Setting}

We applied our method to train on the Deepseek-R1-Distill model as well as the Qwen3-1.7b, Qwen3-8b, and Qwen3-14b models. We performed dual-stage fine-tuning using the LoRA method on our constructed Datasets1 and Datasets2 datasets, with a rank of 32, a learning rate of $2\times10^{-5}$, and 3 epochs of tuning in each stage. During the reinforcement learning phase, we continued to use the data from the SFT stage and additionally incorporated the OpenR1-MATH dataset to enhance the model's exploration space and policy diversity under complex reasoning tasks. The training context was set to 16K, batch size to 64, learning rate to $5\times10^{-6}$, and training was conducted for 1 epoch. By training with these hyperparameters, the model is able to master the backward reasoning mode to solve problems.

We evaluated our method on several classic mathematical reasoning tasks, including three public test sets: GSM8K, MATH-500, and AIME (2024 and 2025), covering a range of problem difficulties, as well as on a self-constructed test set BQ. The BQ dataset comprises 300 backward reasoning problems, with equal proportions (1:1:1) of condition-swap, reverse-intermediate, and error-diagnosis tasks.

We set the model's maximum response length to 16K and the temperature to 1. For automated evaluation, we required the model to enclose the final answer within \texttt{Boxed\{\}}. For the evaluation metrics, we recorded the model's response length and computed accuracy by checking whether the answer within \texttt{Boxed\{\}} matched the ground truth. For samples where the model failed to place the answer in \texttt{Boxed\{\}}, we additionally extracted the last 500 tokens of the model's response (which usually contain the final conclusion) and input them along with the ground truth into Deepseek-V3, restricting the output to 1 or 0 to indicate agreement or disagreement, thereby reducing errors caused by incorrect answer formatting.

\subsection{Baseline}

We compared baseline models with parameter scales ranging from 1b to 32b.

Baseline models of approximately 1B parameters include: (1) Deepseek-R1-Distill-1.5b: a 1.5B-parameter distilled version based on the Deepseek-R1 architecture. (2) qwen2.5-math-1.5b: a 1.5B-parameter model developed by Alibaba, optimized using mathematical datasets. (3) qwen3-1.7b: a newly open-sourced 1.7B-parameter model by Alibaba, supporting both thinking and non-thinking modes; here, we use the thinking mode for evaluation. (4) Llama-3.2-1b: a 1B-parameter lightweight model open-sourced by Meta. (5) Llama-3.2-3b: a 3B-parameter lightweight model open-sourced by Meta. (6) qwen2.5-3b: a 3B-parameter model open-sourced by Alibaba.

Baseline models of approximately 8B parameters include: (1) Deepseek-R1-Distill-7b: a 7B-parameter distilled version based on the Deepseek-R1 architecture. (2) qwen3-8b: an 8B-parameter model open-sourced by Alibaba, evaluated in reasoning mode. (3) GLM-Z1-9b: a 9B-parameter reasoning model open-sourced by Zhipu AI.

Baseline models of 14B parameters and above include: (1) Deepseek-R1-Distill-14b: a 14B-parameter distilled version based on the Deepseek-R1 architecture. (2) qwen3-14b: a 14B-parameter model open-sourced by Alibaba, evaluated in reasoning mode. (3) Deepseek-R1-Distill-32b: a 32B-parameter distilled version based on the Deepseek-R1 architecture.

For the qwen2.5 series and Llama-3.2 series models, we additionally appended the prompt ``Let's think step by step, and place the answer in Boxed\{\}'' during testing. For the other models, we only added the prompt ``Place the answer in Boxed\{\}.''

\subsection{Compared with baseline models}

As shown in Table~\ref{tab:main}, the comparison results across the four benchmark datasets demonstrate that our method achieves consistently strong average performance. Compared with models of approximately 1.5B parameters, our approach yields at least a 1.8\% improvement in average accuracy, indicating that incorporating backward reasoning effectively enhances the model's reasoning capability. Furthermore, relative to long chain-of-thought models such as R1-Distill-1.5B and Qwen3-1.7B, our method not only improves average accuracy but also significantly reduces the average response length, thereby enhancing reasoning efficiency. In the comparison among models with 8B parameters and above, our model again achieves the highest average accuracy while maintaining the shortest average response length, highlighting its superiority in both reasoning quality and efficiency across varying scales.

To further verify the advantages of our method in solving backward reasoning problems, we categorize the problems in GSM8K, MATH-500, and AIME into forward and backward types, and construct three subsets consisting exclusively of backward problems: GSM8K-B, MATH-500-B, and AIME-B. The evaluation results on these subsets are presented in Table~\ref{tab:backward}. Among the 1.5B-scale long-chain-of-thought models, Ours-Qwen3 achieves the highest average accuracy across all three datasets, while Ours-R1 produces the shortest average response length. For the 8B-scale models, although Ours-Qwen3 attains slightly lower accuracy than the 9B GLM-Z1 model on GSM8K-B, it still demonstrates the best overall average accuracy and maintains a comparable average response length. In the comparison of models with 14B parameters and above, our 14B model achieves competitive performance even against the significantly larger Deepseek-R1-Distill-32B.

\begin{table*}[t]
  \centering
  \caption{Performance comparison of various models across the four test datasets.}
  \label{tab:main}
  \small
  \begin{tabular}{l*{10}{c}}
    \toprule
    \multirow{2}{*}{\textbf{Method}} & \multicolumn{2}{c}{\textbf{GSM8K}} & \multicolumn{2}{c}{\textbf{MATH-500}} & \multicolumn{2}{c}{\textbf{AIME-2024}} & \multicolumn{2}{c}{\textbf{BQ}} & \multicolumn{2}{c}{\textbf{Avg}} \\
    \cmidrule(lr){2-3}\cmidrule(lr){4-5}\cmidrule(lr){6-7}\cmidrule(lr){8-9}\cmidrule(lr){10-11}
     & Acc & Length & Acc & Length & Acc & Length & Acc & Length & Acc & Length \\
    \midrule
    \multicolumn{11}{l}{\textit{Short-Cot}} \\
    Qwen2.5-math & 84.8 & 407 & 74.4 & 682 & 16.6 & 938 & 71.0 & 812 & 61.7 & 710 \\
    Qwen2.5-3b & 85.0 & 316 & 62.0 & 608 & 10.0 & 1161 & 57.3 & 1056 & 53.6 & 785 \\
    Llama-3.2-1b & 44.4 & 168 & 18.8 & 686 & 3.3 & 1004 & 19.3 & 1025 & 21.5 & 721 \\
    Llama-3.2-3b & 77.7 & 205 & 29.0 & 657 & 1.6 & 1078 & 30.0 & 1023 & 34.6 & 741 \\
    \addlinespace
    \multicolumn{11}{l}{\textit{Long-Cot-1b}} \\
    R1-Distill-1.5b & 79.0 & 978 & 80.6 & 4887 & 23.3 & 11985 & 75.3 & 5124 & 64.6 & 5744 \\
    Qwen3-1.7b & 90.7 & 1973 & 88.2 & 5001 & 33.3 & 13225 & 79.3 & 5846 & 72.9 & 6511 \\
    Ours-R1-1.5b & 83.7 & 1422 & 84.8 & 2938 & 33.3 & 9730 & 82.3 & 3248 & 71.0 & 4335 \\
    Ours-Q3-1.7b & 90.2 & 1247 & 89.8 & 3915 & 35.0 & 11042 & 83.7 & 4556 & 74.7 & 5190 \\
    \addlinespace
    \multicolumn{11}{l}{\textit{Long-Cot-8b}} \\
    R1-Distill-7b & 88.9 & 1010 & 89.6 & 4500 & 48.3 & 11137 & 86.7 & 4678 & 78.2 & 5314 \\
    Qwen3-8b & 95.9 & 2316 & 89.6 & 5743 & 56.7 & 12324 & 88.3 & 5902 & 82.7 & 6571 \\
    GLM-Z1-9b & 95.0 & 979 & 92.0 & 3375 & 60.0 & 9203 & 90.3 & 3742 & 84.3 & 4325 \\
    Ours-Q3-8b & 96.8 & 924 & 92.0 & 2896 & 58.3 & 9578 & 91.7 & 3024 & 84.7 & 4106 \\
    \addlinespace
    \multicolumn{11}{l}{\textit{Long-Cot-14b}} \\
    R1-Distill-14b & 93.6 & 1006 & 91.6 & 3614 & 51.7 & 10263 & 88.7 & 3766 & 81.4 & 4662 \\
    Qwen3-14b & 96.4 & 1870 & 93.6 & 4621 & 66.7 & 12058 & 90.3 & 4852 & 87.4 & 5850 \\
    R1-Distill-32b & 94.8 & 782 & 92.6 & 3525 & 65.0 & 10500 & 91.3 & 3510 & 85.9 & 4579 \\
    Ours-Q3-14b & 97.2 & 987 & 94.8 & 2795 & 66.7 & 9421 & 93.0 & 3022 & 87.9 & 4056 \\
    \bottomrule
  \end{tabular}
\end{table*}

\begin{table*}[t]
  \centering
  \caption{Performance comparison of various models on the backward reasoning subsets across the four test datasets.}
  \label{tab:backward}
  \small
  \begin{tabular}{l*{8}{c}}
    \toprule
    \multirow{2}{*}{\textbf{Method}} & \multicolumn{2}{c}{\textbf{GSM8K-B}} & \multicolumn{2}{c}{\textbf{MATH-500-B}} & \multicolumn{2}{c}{\textbf{AIME-B}} & \multicolumn{2}{c}{\textbf{Avg}} \\
    \cmidrule(lr){2-3}\cmidrule(lr){4-5}\cmidrule(lr){6-7}\cmidrule(lr){8-9}
     & Acc & Length & Acc & Length & Acc & Length & Acc & Length \\
    \midrule
    \multicolumn{9}{l}{\textit{Short-Cot}} \\
    Qwen2.5-math & 78.5 & 397 & 57.5 & 894 & 25.0 & 1018 & 53.7 & 770 \\
    Qwen2.5-3b & 75.9 & 335 & 42.5 & 779 & 8.3 & 820 & 42.2 & 645 \\
    Llama-3.2-1b & 11.4 & 189 & 12.5 & 798 & 0 & 1390 & 8.0 & 792 \\
    Llama-3.2-3b & 59.5 & 223 & 15.0 & 1247 & 0 & 765 & 24.8 & 745 \\
    \addlinespace
    \multicolumn{9}{l}{\textit{Long-Cot-1b}} \\
    R1-Distill-1.5b & 65.2 & 1401 & 73.8 & 6313 & 25.0 & 12993 & 54.7 & 6902 \\
    Qwen3-1.7b & 84.8 & 2650 & 80.0 & 7268 & 25.0 & 13760 & 63.3 & 7893 \\
    Ours-R1-1.5b & 86.0 & 1167 & 76.3 & 4289 & 33.3 & 10845 & 65.2 & 5034 \\
    Ours-Q3-1.7b & 89.9 & 1867 & 82.5 & 5358 & 33.3 & 11906 & 68.6 & 6377 \\
    \addlinespace
    \multicolumn{9}{l}{\textit{Long-Cot-8b}} \\
    R1-Distill-7b & 77.2 & 1667 & 82.5 & 5564 & 33.3 & 11803 & 64.3 & 6345 \\
    Qwen3-8b & 92.4 & 2598 & 89.4 & 7101 & 50.0 & 12170 & 77.3 & 7290 \\
    GLM-Z1-9b & 96.3 & 1291 & 87.5 & 4524 & 66.7 & 9234 & 83.5 & 5016 \\
    Ours-Q3-8b & 94.9 & 1076 & 92.5 & 4166 & 66.7 & 10283 & 84.7 & 5175 \\
    \addlinespace
    \multicolumn{9}{l}{\textit{Long-Cot-14b}} \\
    R1-Distill-14b & 84.8 & 1507 & 83.8 & 5334 & 58.3 & 9919 & 75.6 & 5587 \\
    Qwen3-14b & 92.4 & 2164 & 91.3 & 6384 & 58.3 & 12395 & 80.7 & 6981 \\
    R1-Distill-32b & 88.9 & 1150 & 85.7 & 4775 & 58.3 & 11397 & 77.6 & 5774 \\
    Ours-Q3-14b & 94.9 & 1154 & 93.8 & 3829 & 66.7 & 9816 & 85.1 & 4933 \\
    \bottomrule
  \end{tabular}
\end{table*}

\begin{table*}[t]
  \centering
  \caption{Comparison of model performance across training stages.}
  \label{tab:ablation}
  \small
  \begin{tabular}{l*{10}{c}}
    \toprule
    \multirow{2}{*}{\textbf{Method}} & \multicolumn{2}{c}{\textbf{GSM8K}} & \multicolumn{2}{c}{\textbf{MATH-500}} & \multicolumn{2}{c}{\textbf{AIME}} & \multicolumn{2}{c}{\textbf{BQ}} & \multicolumn{2}{c}{\textbf{Avg}} \\
    \cmidrule(lr){2-3}\cmidrule(lr){4-5}\cmidrule(lr){6-7}\cmidrule(lr){8-9}\cmidrule(lr){10-11}
     & Acc & Length & Acc & Length & Acc & Length & Acc & Length & Acc & Length \\
    \midrule
    R1-Distill-1.5b & 79.0 & 978 & 80.6 & 4887 & 23.3 & 11985 & 75.3 & 5124 & 64.6 & 5746 \\
    +SFT stage1+2 & 80.3 & 1064 & 81.4 & 4279 & 25.0 & 12717 & 76.7 & 4979 & 65.9 & 5851 \\
    +SFT stage1 & 81.0 & 1426 & 81.8 & 4909 & 25.0 & 12165 & 77.0 & 4953 & 66.2 & 5863 \\
    +SFT stage2 & 80.9 & 1674 & 83.0 & 4452 & 30.0 & 12974 & 79.0 & 4512 & 68.2 & 5903 \\
    +RL & 83.7 & 1422 & 84.8 & 2938 & 33.3 & 9730 & 82.3 & 3248 & 71.0 & 4335 \\
    \addlinespace
    Qwen3-1.7b & 90.7 & 1973 & 88.2 & 5001 & 33.3 & 13225 & 79.3 & 5846 & 72.9 & 6511 \\
    +SFT stage1+2 & 88.4 & 1796 & 88.0 & 5517 & 31.7 & 13187 & 80.3 & 5804 & 72.1 & 6576 \\
    +SFT stage1 & 89.3 & 1593 & 87.6 & 5138 & 31.7 & 13137 & 80.0 & 5645 & 72.2 & 6378 \\
    +SFT stage2 & 88.6 & 1842 & 88.4 & 5176 & 35.0 & 13391 & 81.7 & 5912 & 73.4 & 6580 \\
    +RL & 90.2 & 1247 & 89.8 & 3915 & 35.0 & 11042 & 83.7 & 4556 & 74.7 & 5190 \\
    \addlinespace
    Qwen3-8b & 95.9 & 2316 & 89.8 & 5743 & 56.7 & 12324 & 88.3 & 5902 & 82.7 & 6571 \\
    +SFT stage1+2 & 96.2 & 1275 & 90.2 & 4133 & 51.7 & 11660 & 88.7 & 4301 & 81.7 & 5342 \\
    +SFT stage1 & 96.8 & 1359 & 90.4 & 4226 & 48.3 & 11597 & 89.0 & 4320 & 81.1 & 5376 \\
    +SFT stage2 & 95.4 & 1236 & 88.6 & 3626 & 55.0 & 11145 & 89.0 & 4421 & 82.0 & 5107 \\
    +RL & 96.8 & 924 & 92.0 & 2896 & 58.3 & 9578 & 91.7 & 3024 & 84.7 & 4106 \\
    \addlinespace
    Qwen3-14b & 96.4 & 1870 & 93.6 & 4621 & 66.7 & 12058 & 90.3 & 4852 & 87.4 & 5850 \\
    +SFT stage1+2 & 96.2 & 1702 & 94.2 & 3854 & 58.3 & 10692 & 91.0 & 4532 & 84.9 & 5195 \\
    +SFT stage1 & 96.8 & 1469 & 94.4 & 3536 & 55.0 & 10547 & 90.3 & 4415 & 84.7 & 4992 \\
    +SFT stage2 & 96.4 & 1401 & 94.0 & 3450 & 61.7 & 10966 & 91.7 & 4578 & 86.0 & 5099 \\
    +RL & 97.2 & 987 & 94.8 & 2795 & 66.7 & 9421 & 93.0 & 3022 & 87.9 & 4056 \\
    \bottomrule
  \end{tabular}
\end{table*}

\begin{figure}[t]
  \centering
  \includegraphics[width=0.75\linewidth]{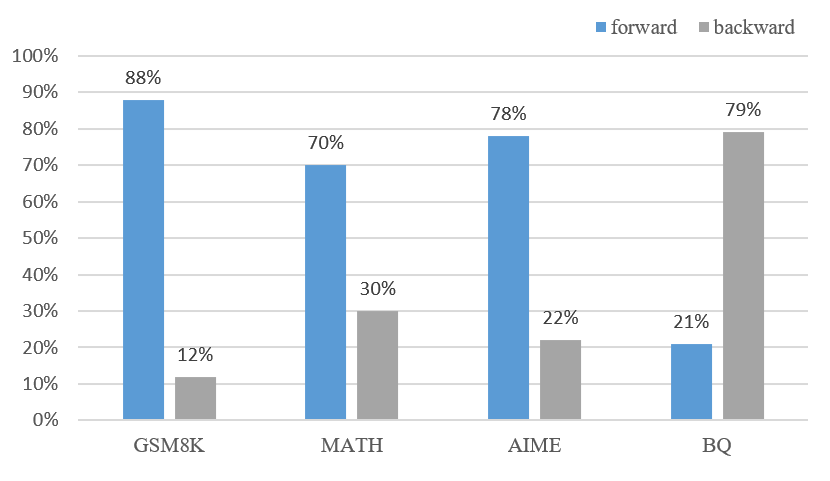}
  \caption{Comparison of forward and backward reasoning ratios.}
  \label{fig:ratio}
\end{figure}

For the Ours-Q3-14b model, which achieved the highest average accuracy, we also calculated the proportion of forward and backward reasoning used across four test datasets. As shown in Fig.~\ref{fig:ratio}. For GSM8K, MATH-500, and AIME, the proportions of backward reasoning are 12\%, 30\%, and 22\%, respectively, whereas for the BQ dataset, the proportion reaches 79\%, since BQ mainly consists of backward reasoning problems. These findings indicate that our method successfully enables the model to autonomously choose backward reasoning when appropriate.

\subsection{Ablation Experiment}

We evaluated the results of each training stage of the model, and the results are shown in Table~\ref{tab:ablation}. Here, SFT stage1+2 denotes performing a single supervised fine-tuning on the baseline model using the combined Datasets1 and Datasets2; SFT stage1 refers to applying one supervised fine-tuning on the baseline model using Datasets1; SFT stage2 represents performing another supervised fine-tuning using Datasets2 on top of SFT stage1; and RL indicates reinforcement learning training based on SFT stage2. All training parameters are consistent with those described in the experimental setting section.

Compared with the dual-stage supervised fine-tuning, the single supervised fine-tuning performed on the merged dataset yields a smaller improvement in average accuracy than that achieved by SFT stage2. For the 1.5B-parameter model, the supervised fine-tuning stages (Stage 1 \& Stage 2) improve its performance across datasets of varying difficulty, but they also increase the model's response length to some extent, and the proportion of backward reasoning used remains low. For example, in the first supervised fine-tuning stage, the R1-Distill-1.5B model achieves a 1.6 points improvement in average test accuracy, but its average reasoning length increases by 2\%. The second stage of supervised fine-tuning further increases the average accuracy by 2 points, but the response length becomes 2.7\% longer compared with the initial model. After reinforcement learning training, the model's performance on the four types of datasets is further improved, the average response length decreases by 24.6\%, and for the BQ test set, which consists of backward-reasoning problems, the proportion of backward reasoning in the model's generation increases significantly. We attribute these effects to the fine-grained reward function and the balanced sampling strategy. For the 8B and 14B parameter qwen3 models, SFT Stage 1 significantly improves their performance on the GSM8K and MATH test sets but leads to performance drops on the more challenging datasets. In contrast, SFT Stage 2 mainly enhances performance on AIME, which is related to the distinct emphases of the training data used in the two stages. After reinforcement learning, our models achieve average accuracy improvements of 2 points and 0.5 points over the initial models, while their average response lengths are reduced by 37.5\% and 30.7\%, respectively.

\section{Conclusion}

In this work, we introduce a construction approach for a backward reasoning mode, aimed at strengthening the model's ability to engage in backward reasoning and improving its adaptive behavior in complex reasoning scenarios. Unlike baseline models that rely exclusively on forward reasoning, our backward reasoning model can dynamically choose appropriate reasoning strategies depending on the problem type. Experimental results demonstrate that the proposed method significantly improves the model's reasoning performance across multiple mathematical reasoning benchmarks (including GSM8K, MATH-500, AIME and BQ), while effectively reducing response length overhead during inference. These findings not only verify the importance of backward reasoning in complex reasoning tasks but also offer a novel and effective pathway for further enhancing model reasoning capabilities. In the future, we plan to extend the application potential of the backward reasoning mode to other domains (such as scientific reasoning and program verification) to validate its cross-domain generalizability. In addition, we aim to further explore other reasoning modes, such as bidirectional reasoning, to assess their effectiveness in enhancing large language model reasoning, where reasoning progresses simultaneously from both the conditions and the goals until they meet in the middle.


\end{document}